\documentclass[runningheads]{llncs}

\usepackage[width=122mm,left=12mm,paperwidth=146mm,height=193mm,top=12mm,paperheight=217mm]{geometry}

\usepackage[T1]{fontenc}
\usepackage{graphicx}
\usepackage{booktabs}
\usepackage[misc]{ifsym}

\usepackage{mwe}
\usepackage{verbatim}
\usepackage{multirow}
\usepackage[normalem]{ulem}
\usepackage{enumitem}
\usepackage[ruled, vlined, linesnumbered]{algorithm2e}
\usepackage{algorithmicx, algpseudocode}

\newcommand\red[1]{\textcolor{red}{#1}}

\usepackage{amssymb}
\usepackage{amsmath} 
\usepackage{subcaption}
\usepackage[table]{xcolor}
\usepackage{xurl}

\newcommand{\method}{SLeDGe\xspace}
\newcommand{\methodlight}{SLeDGe-L\xspace}

\begin{document}

\title{SLeDGe: Semi-Supervised Learning on Data Streams with Graph Structure Learning}

\titlerunning{Semi-Supervised Learning on Data Streams with Graph Structure Learning}

\author{Heechan Moon\inst{1} \and
Kijung Shin\inst{1} (\Letter)}
\authorrunning{H. Moon and K. Shin}


\institute{KAIST, Seoul, Republic of Korea \\
\email{\{heechan9801,kijungs\}@kaist.ac.kr}}

\toctitle{SLeDGe: Semi-Supervised Learning on Data Streams with Graph Structure Learning}
\tocauthor{Heechan Moon, Kijung Shin}

\maketitle              

\begin{abstract}

Semi-supervised learning (SSL) on data streams is challenging due to the continuous evolution of high-volume data and the scarcity of labels.
Existing methods are limited in leveraging the intrinsic relationships among samples because they typically rely on fixed similarity measures or static graph structures, which cannot capture how relationships evolve over time.
We propose \method, an SSL method for data streams that jointly learns a predictive model and an adaptive graph structure under strict memory and label constraints.
\method maintains compact labeled and unlabeled memories using distinct update strategies, balancing rapid adaptation to novel features with the retention of historical consistency.
In addition, by encouraging sparsity in the relational graph, \method filters out spurious connections and enables effective propagation of label supervision.
Across 12 datasets, \method outperforms state-of-the-art competitors, achieving average relative accuracy gains of 31.7\% with 0.1\% labels and 14.8\% with 1\% labels.

\end{abstract}

\section{Introduction}

Learning from data streams poses a significant challenge in modern machine learning, particularly when labeled data are scarce.
In many web applications, such as webpage classification~\cite{cmu_webkb4unis_1998} and news classification~\cite{petukhova2023mn}, data arrive continuously and in high volume.
For real-time processing, samples must be maintained in memory; however, the cumulative stream quickly exceeds available system resources.
At the same time, obtaining labels is often costly or impractical.

To address these challenges, semi-supervised learning (SSL) on data streams has received increasing attention, as it enables learning from a small number of labels while adapting to a continuously evolving flow of information~\cite{din2020online,wagner2018semi}. Notably, general SSL techniques~\cite{van2020survey,yang2022survey} are mostly unsuitable in streaming settings because they assume static, offline datasets or rely on iterative training passes over the entire data corpus to refine pseudo-labels.

However, existing techniques for SSL on data streams~\cite{din2021learning,khezri2020stds,le2019semi} often have limited ability to exploit inherent relationships among samples.
They either treat samples independently or rely on fixed similarity heuristics, preventing them from modeling how relationships evolve over time.
As a result, label supervision cannot be effectively propagated over samples.
While graph structure learning (GSL) offers a promising way to effectively learn the relation between samples, most GSL methods are designed for offline settings and are unsuitable for streaming scenarios with strict memory limits and severe label scarcity.

To fill this gap, we propose \textbf{\method} (\textbf{\underline{S}}emi-supervised \textbf{\underline{Le}}arning on \textbf{\underline{D}}ata Stream with \textbf{\underline{G}}raph structur\textbf{\underline{e}} learning), which jointly learns a predictive model and an adaptive graph structure under scarce labels and limited storage. 
Instead of relying on a fixed graph or heuristic similarity measures, \method integrates dynamic GSL into streaming settings, capturing evolving inter-sample relationships. 
\method retains two compact memories: a \textit{labeled memory} optimized for plasticity to rapidly integrate novel features, and an \textit{unlabeled memory} focused on stability to preserve historical consistency.
\method updates a graph over both memories and the current sample to propagate label supervision.

The strengths of \method are summarized as follows:
\begin{itemize}[leftmargin=*]
\item \textbf{Adaptive.} \method learns a dynamic graph structure that captures evolving inter-sample dependencies, instead of relying on fixed similarity metrics. The graph enables effective propagation of label supervision across samples.
\item \textbf{Efficient.} By enabling GSL over compact memories, \method remains reliable even under extreme label scarcity and tight memory budgets.
\item \textbf{Effective.} 
Across 12 datasets, \method outperforms state-of-the-art competitors with average accuracy gains of 31.7\% with 0.1\% labels and 14.8\% with 1\% labels.
We further confirm the advantage of dynamic GSL over static ones.
\end{itemize}
For reproducibility, we provide our code and datasets at 
\url{https://github.com/Heechan-Moon/SLeDGe}.

\section{Related Work}

\subsection{Semi-Supervised Learning on Data Streams}
Semi-supervised learning (SSL) on data streams has gained increasing attention due to its relevance in large-scale, evolving data scenarios with limited labels.
Graph-based methods~\cite{wagner2018semi} update graph structures for temporal label propagation.
Cluster-based approaches~\cite{din2021learning,din2020online} update cluster representations and assign pseudo-labels using cluster proximity. 
Other works~\cite{le2019semi,khezri2020stds} employ self-training or ensemble strategies. 
Recent studies have further expanded on these approaches by introducing synchronization techniques to handle temporal shifts on data streams~\cite{din2024synchronization}, adapting to weakly labeled  streams~\cite{zhang2022adaptive}, and extending label-efficient continual learning to complex domains like video~\cite{wu2023label}.
However, these methods either ignore dependencies among samples or rely on fixed similarity measures that fail to capture evolving relationships.
Our method, in contrast, learns and continually adapts the relational structure as the stream evolves.

\subsection{Graph Structure Learning}
Instead of relying on fixed graph structures, graph structure learning (GSL) aims to jointly optimize graph topology and node representations by refining an existing graph~\cite{shen2024unsupervised,han2025uncertainty,liu2022towards} or inferring connectivity from attributes~\cite{kalofolias2016learn,liu2022towards}.
However, most GSL methods operate offline, making them unsuitable for streaming environments with strict memory and label constraints.
Our framework fills this gap by adaptively generating a graph structure based on compact memories.

\section{Problem Description}

We consider streaming semi-supervised learning 
where data arrive sequentially as $\mathcal{D} = \{x_t\}_{t=1}^{T}$, with each sample $x_t \in \mathbb{R}^d$. 
Only a small subset of samples has observed labels $y_t \in C$, where $C$ is the set of classes,
and no separate training set is provided beyond the stream.
The learner must update the model online while operating under strict memory constraints, 
with a fixed total memory capacity $M$.
The objective is to classify each instance $x_t$ while adapting to an evolving data stream under severe label scarcity and strict storage constraints.
Table~\ref{tab:notations} summarizes key notations.

\begin{table}[t]
\centering
\caption{Summary of frequently used notations.}
\begin{tabular}{l|l}
\toprule
\textbf{Notation} & \textbf{Description} \\
\midrule
$x_t, y_t$ & Input sample and its label at time $t$ \\
$d$ & Dimension of input sample \\
$M_L, M_U$ ($L, U$) & Labeled/unlabeled memory (and their sizes) \\
$M$ & Total memory size  $(= L + U)$ \\
$f_E, f_S, f_P, f_B$ & Embedding, scoring, pruning, and classification functions \\
$X_t, A_t$ & Input feature and sparse adjacency matrix at time $t$ \\
$k, h$ & Number of neighbors and embedding dimension \\
$\tau$ & Temporal decay parameter \\
\bottomrule
\end{tabular}
\label{tab:notations}
\end{table}

\section{Proposed Method}

\begin{figure}[t!]
    \centering
    \begin{subfigure}{\textwidth}
        \centering
        \includegraphics[width=\textwidth]{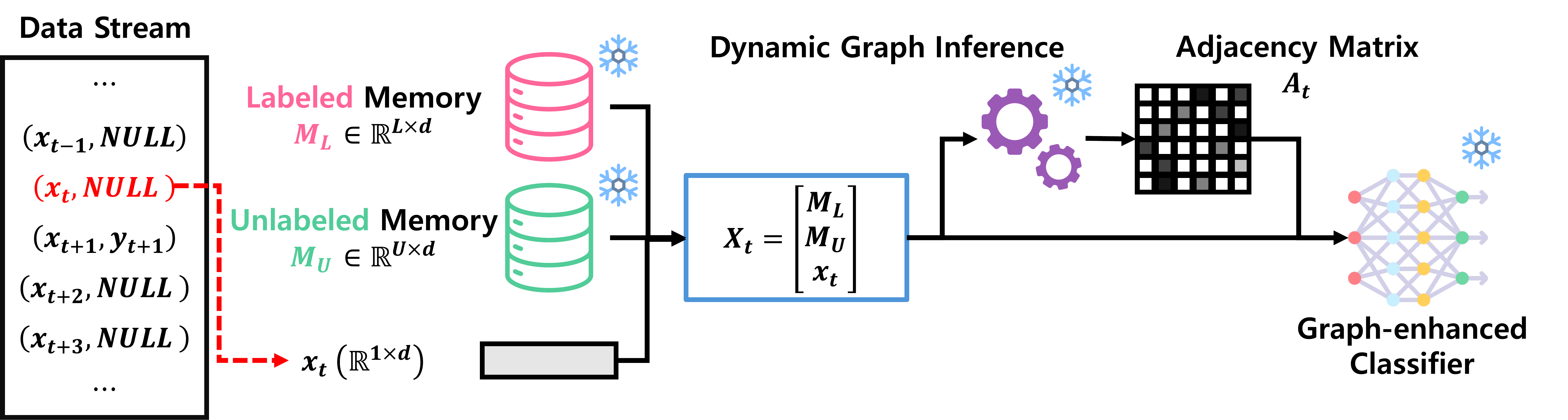}
        \caption{Graph-enhanced Classification via Dynamic Graph Inference (Section~\ref{method:GSL})}
        \label{fig:sledge_infer}
    \end{subfigure}
    
    \begin{subfigure}{\textwidth}
        \centering
        \includegraphics[width=\textwidth]{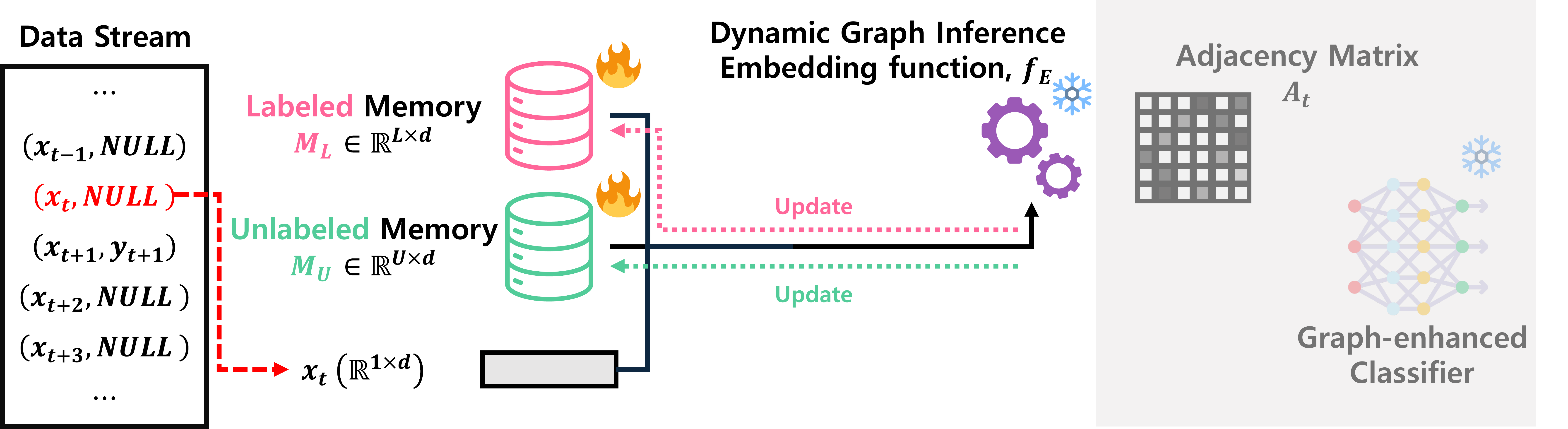}
        \caption{Memory Update (Section~\ref{method:memory})}
        \label{fig:sledge_memory_update}
    \end{subfigure}

    \begin{subfigure}{\textwidth}
        \centering
        \includegraphics[width=\textwidth]{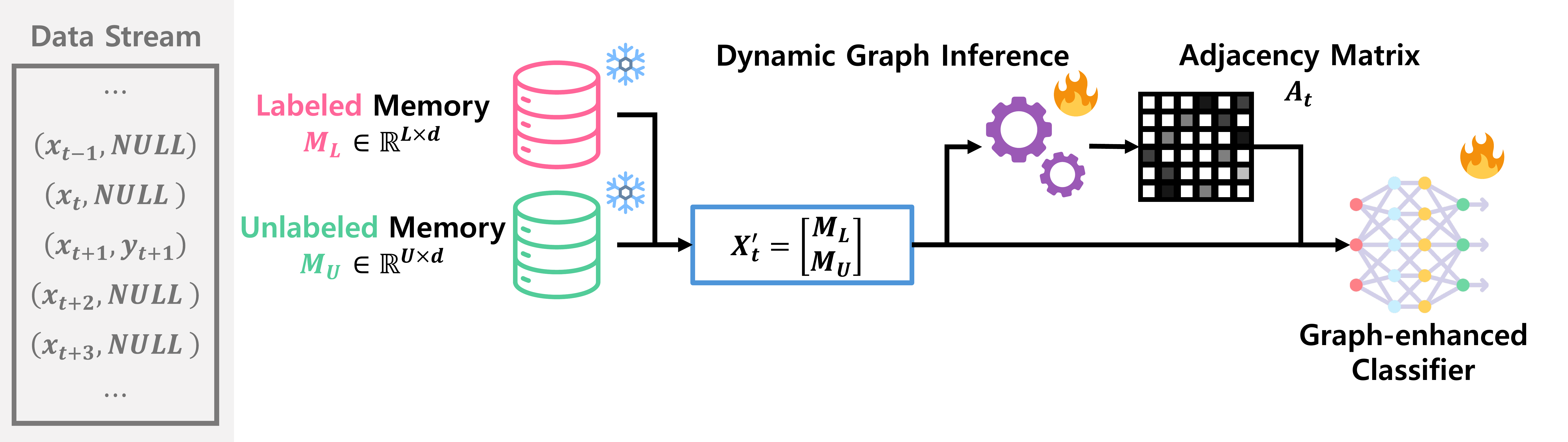}
        \caption{Model Update (Section~\ref{method:training})}
        \label{fig:sledge_model_update}
    \end{subfigure}
    
    \caption{ \label{fig:overview}
        Overview of \method.
        \textit{(a) Graph-enhanced Classification via Dynamic Graph Inference:} Classifying a new sample $x_t$ via dynamic graph inference using memory buffers. (b) \textit{Memory Update}: Updating the buffers with $x_t$ using the embedding function $f_E$. (c) \textit{Model Update}: Refining the classifier and inference modules based on the updated memory state.
    }
\end{figure}

In this section, we introduce \textbf{\method} (\textbf{\underline{S}}emi-supervised \textbf{\underline{Le}}arning on \textbf{\underline{D}}ata Stream with \textbf{\underline{G}}raph structur\textbf{\underline{e}} learning), a framework designed for 
semi-supervised learning on data streams.
As illustrated in Figure~\ref{fig:overview}, for each arriving sample $x_t$, \method proceeds in three stages:
(a) graph-enhanced classification for a newly arrived sample $x_t$ via dynamic graph inference using labeled and unlabeled memory buffers;  
(b) memory update, where buffers are updated with $x_t$ using the embedding function $f_E$; 
and (c) model update, in which the model is optimized using the updated graph structure.
The details of these components are provided in the subsequent subsections.
We then provide a complexity analysis, followed by the introduction of \methodlight, a lightweight variant of \method.

\subsection{Memory Update}\label{method:memory}
A major challenge for models operating on data streams is the need to continuously adapt to new inputs while adhering to strict storage constraints.
Since retaining all past data in main memory is infeasible in practice, 
\method maintains two types of memory, \textit{labeled memory} and \textit{unlabeled memory}, which selectively store informative labeled and unlabeled samples, respectively.

\subsubsection{Labeled Memory ($M_L$).}\label{method:labeled_memory}

The labeled memory, denoted as $M_L$, is dynamically updated with incoming labeled data. 
Given a total capacity $L$ and a set of classes $C$ where $|C| = c$, the memory is uniformly partitioned, allocating a fixed slot size of $\lfloor L/c \rfloor$ to each class. 
Upon the arrival of a new labeled sample, \method evaluates the available capacity for its corresponding class. 
If the allocated partition has not reached its limit, the sample is inserted as a new prototype.

Once a class partition reaches capacity, \method updates the most relevant prototype within that class.
The target prototype $p_l$ is selected from the memory subset belonging to class $y_t$ (denoted as $M_L^{(y_t)}$) based on cosine similarity ($\text{sim}$) in the latent space, using the embedding function $f_E$ (defined in Section~\ref{method:GSL_embedding}):  
\begin{equation} p_l = \arg\max\nolimits_{p_{l_i} \in M_L^{(y_t)}} \text{sim}(f_E(x_t), f_E(p_{l_i})).\end{equation}
To balance stability and plasticity, the selected prototype $p_l$ is updated via a weighted combination of its current state and the new input $x_t$. \method assigns a weight $\alpha$ to $x_t$ to facilitate the proactive integration of novel features: \begin{equation} p_l' \leftarrow (1 - \alpha) p_l + \alpha x_t.\end{equation}
The coefficient $\alpha= \left(1 - \frac{1}{N}\right) \text{sim}(f_E(x_t), f_E(p_l))$ is determined by the similarity between the embeddings and the update frequency $N$ of the selected prototype.
In practice, $\alpha$ tends to be large, giving more weight to $x_t$ and encouraging faster adaptation.
Finally, the prototype is updated following $L_2$ normalization: \begin{equation}p_l \gets p_l' / \|p_l'\|_2.\end{equation}

\subsubsection{Unlabeled Memory ($M_U$).}\label{method:unlabeled_memory}
The unlabeled memory, $M_U$, is continuously updated with incoming unlabeled inputs. 
If the memory retains available capacity, the sample is inserted as a new, distinct prototype. 
Otherwise, as in the labeled memory update process, \method identifies the most relevant prototype: \begin{equation}p_u = \arg\max\nolimits_{p_{u_j} \in M_U} \mathrm{sim}(f_E(x_t), f_E(p_{u_j})).\end{equation}
The state of the selected prototype is then updated as follows: \begin{equation}p_u' \leftarrow \beta p_u + (1-\beta) x_t,\end{equation}
where $\beta = \left(1 - \frac{1}{N}\right) \text{sim}(f_E(x_t), f_E(p_u))$.
In practice, $\beta$ tends to be large, giving more weight to the prior prototype $p_u$. 
That is, the unlabeled memory favors the gradual accumulation of information, reflecting the uncertainty in unlabeled samples.
This contrasts with the labeled memory, which prioritizes rapid adaptation to $x_t$.
Finally, the selected prototype is updated following $L_2$ normalization: 
\begin{equation}p_u \gets p_u' / \|p_u'\|_2.\end{equation}

These mechanisms enable the prototypes to dynamically evolve and accurately reflect their respective regions in the latent space while maintaining stable magnitudes. 
By employing dual, tailored update strategies that prioritize plasticity in the labeled memory to rapidly integrate novel features and emphasize stability in the unlabeled memory to preserve historical consistency, \method effectively balances continuous adaptation with the retention of past knowledge.

\subsection{Graph-enhanced Classification via Dynamic Graph Inference}\label{method:GSL}
\method adaptively constructs a graph to capture evolving dependencies among stored samples and the current input.
The learner $f$ consists of (1) an \textit{embedding function} ($f_E$), (2) a \textit{scoring function} ($f_S$), and (3) a \textit{pruning function} ($f_P$).

\subsubsection{Embedding Function ($f_E$).}\label{method:GSL_embedding}

The graph inference process begins with the input $X_t$, a concatenation of the labeled memory, the unlabeled memory, and the current input sample at time $t$, i.e.,
\begin{equation}
    X_t=[M_L; M_U; x_t] \in \mathbb{R}^{(M + 1) \times d},
\end{equation} where $M=L+U$ denotes the total number of stored memory elements.
Given the input feature $X_t \in \mathbb{R}^{(M + 1) \times d}$,
the embedding function $f_E$ maps each row of $X_t$ to a latent space, i.e.,  $\forall i\in \{1,\cdots, (M + 1)\}$:
\begin{equation}
E_t(i) = f_{E}(X_t(i))  \in \mathbb{R}^{h}, \quad f_E : \mathbb{R}^{d} \rightarrow \mathbb{R}^{h}.
\end{equation}
We use an $\alpha_1$-layer MLP as the function $f_E$.

\subsubsection{Scoring Function ($f_S$).}\label{method:GSL_scoring}

Given the embeddings $E_t \in \mathbb{R}^{(M + 1) \times h}$, the scoring function $f_S$ quantifies pairwise relational weights, giving a dense adjacency matrix $A_t' \in \mathbb{R}^{(M + 1) \times (M + 1)}$, i.e., $\forall i,j \in \{1,\cdots, (M + 1)\}$,
\begin{equation}\label{method:fs}
\left[A_{t}'\right]_{ij} = f_S(E_t (i), E_t (j)), \quad f_S : \mathbb{R}^h \times \mathbb{R}^h \rightarrow \mathbb{R}.
\end{equation}
We adopt cosine similarity as the scoring function, as it effectively captures the relational structure in a normalized latent space.

\subsubsection{Pruning Function ($f_P$).}\label{method:GSL_pruning}

The dense matrix $A_t'$ is computationally expensive to use directly and may contain spurious or redundant connections. 
To improve efficiency and robustness, we apply a pruning function $f_P$ that retains only the top-$k$ most relevant connections per node:
\begin{equation}
A_t = f_P(A_t', k).
\end{equation}
That is, each row in $A_t$ retains exactly $k$ non-zero values corresponding to the highest-scoring connections, while all other entries are zeroed out.
This operation produces a sparse adjacency matrix $A_t$ that focuses on the most informative relational signals.

\subsubsection{Graph-enhanced Classifier ($f_{B}$).}\label{method:GSL_usage}
The learned adjacency matrix $A_t$ provides the structure for information propagation between samples, thereby enabling supervision to propagate as well. 
At each layer $l\in\{1,\cdots,\alpha_2\}$, 
each row of the hidden representation $H^{(l-1)}$ with $H^{(0)}=X_t$ 
is first passed through a \textit{backbone} model $f^{(l)}_{B}$, which is implemented as a fully connected layer with layer normalization due to its flexibility, i.e., $\forall i\in\{1,\cdots,(M + 1)\}$, 
\begin{equation}
    \bar{H}^{(l)}(i) = f_{B}^{(l)}(H^{(l-1)}(i)). 
\end{equation} 
Then, the resulting representations are refined via graph-based message passing, 
\begin{equation}\label{eq:combine}
H^{(l)} = \hat{A_t} \bar{H}^{(l)},
\end{equation}
where $\hat{A_t}=\tilde{D}^{-1/2}(A_t+I)\tilde{D}^{-1/2}$ is the symmetrically normalized adjacency matrix with self-loops, and $\tilde{D}$ is the degree matrix of $(A_t + I)$.
This normalization stabilizes propagation and prevents high-degree nodes from dominating.

Collectively, the classifier applies  $\alpha_2$-layers of GCN \cite{kipf2016semi}-style updates, with the final-layer representations as logits,\footnote{The backbone model in the final layer sets the output dimension to the class number.} i.e.,
\begin{equation}
pred(X_t, A_t)=softmax(H^{(\alpha_2)}).    \label{eq:pred}
\end{equation}
This design enables the model to exploit both node-level features and the inferred relational structure.
Note that, if the target application uses a domain-specific model (e.g., for image, text, or tabular inputs),
the backbone models above can be replaced accordingly, allowing \method’s learned relational structure to be integrated in the same way.
This allows specialized architectures to leverage graph-based relational reasoning without altering their core design.

\subsection{Model Update}\label{sec:objective}\label{method:training}
From the updated memory, \method constructs $X'_t = [M_L; M_U] \in \mathbb{R}^{M \times d}$ and updates its model parameters (i.e., those in $f_E$ and $f^{(l)}_{B}$'s) via a supervised loss $\mathcal{L}_{L}^{(t)}$ defined over the labeled memory $M_L$.
Specifically, to adequately reflect newly updated samples, 
we apply temporal weighting, assigning exponentially decaying importance to samples based on their age, i.e., 
\begin{equation}\label{eq:labeled}
\mathcal{L}_{L}^{(t)}
= \frac{1}{|M_L|} \sum\nolimits_{(x_i, y_i)\in M_L} w_i \, \mathcal{L}(\hat{y}_i, y_i),
\end{equation}
where $\hat{y}_i$ is the prediction result for $x_i$ in Eq.~\eqref{eq:pred}, $\mathcal{L}$ is a classification loss (spec., cross-entropy), and $w_i = e^{-t_i / \tau}$ is the temporal weight.  
Here, $t_i$ is the temporal index of sample $x_i$, with the newest sample assigned index $0$ and the oldest sample assigned $L-1$, and $\tau$ controls the decay rate.

To stabilize dynamic graph inference, we incorporate a structural regularization term encouraging consistency between pairwise similarities in the input space ($X'_t$) and in the latent space ($E'_t=f_E(X'_t)$):
\begin{equation}\label{eq:reg}
\mathcal{L}_{reg}^{(t)} = \text{MSELoss}\left(\frac{X'_t {X'_t}^\top}{\|X'_t {X'_t}^\top\|_F}, \frac{E'_t {E'_t}^\top}{\|E'_t {E'_t}^\top\|_F}\right)
\end{equation}

The overall training objective combines the supervised loss and the structural regularization term: 
\begin{equation}\label{eq:total_loss}
\mathcal{L}^{(t)} = \mathcal{L}_{L}^{(t)} + \lambda \mathcal{L}_{reg}^{(t)},
\end{equation}
where $\lambda$ is a hyperparameter that controls the balance between the task objective and the regularization.
This combination ensures both stable inference of the graph and effective propagation of supervision over the graph.

\begin{table}[t!]
\centering
\caption{Time and space complexity analysis of \method and \methodlight. * denotes temporary working memory. 
These intermediate tensors are generated and discarded within each processing step and are therefore excluded from the reported persistent memory complexity.}
\begin{tabular}{c|c|c|c}
\toprule
\multicolumn{2}{c|}{\textbf{Step}} & \textbf{Time} & \textbf{Space} \\
\midrule
Memory & $M_L$ & $\mathcal{O}(\frac{L}{c}  h)$ & $\mathcal{O}(L  d)$ \\
Update & $M_U$ & $\mathcal{O}(U  h)$ & $\mathcal{O}(U  d)$ \\
\midrule
  & $f_E$ & $\mathcal{O}(\alpha_1 M  h^2)$ & $\mathcal{O}(\alpha_1  h^2)$, $\mathcal{O}(M  h)^*$\\
Classification & $f_S, f_P$ & $\mathcal{O}(M^2  h), \mathcal{O}(M^2)$  & $\mathcal{O}(M^2)^*$ \\
\& & $f_{S\&P}$ & $\mathcal{O}(c  M  h + c  M )$ & $\mathcal{O}(c  M)^*$ \\
Model & $f_B$  & $\mathcal{O}(\alpha_2  M  h^2)$ & $\mathcal{O}(\alpha_2  h^2)$ \\
Update & $\mathcal{L}_{reg}$ & $\mathcal{O}(M^2  h )$ &  $\mathcal{O}(M^2)^*$ \\
& $\mathcal{L}_{reg}^{L}$ & $\mathcal{O}(c  M  h)$ &  $\mathcal{O}(c  M)^*$ \\
\midrule
\multirow{2}{*}{Total} & \method & $\mathcal{O}(M^2 h + (\alpha_1 + \alpha_2) M h^2)$ & \multirow{2}{*}{$\mathcal{O}(M d + (\alpha_1 + \alpha_2)  h^2)$} \\
 & \method-L & $\mathcal{O}(c  M  h + (\alpha_1 + \alpha_2) M h^2)$ &  \\
\bottomrule
\end{tabular}
\label{tab:complexity}
\end{table}

\begin{table}[t!]
\centering
\caption{Time and space complexity analysis of existing methods.}
\begin{tabular}{c|c|c|c}
\toprule
 & TLP~\cite{wagner2018semi} & ReSSL~\cite{din2020online} & HDSSL~\cite{din2021learning} \\
\midrule
\textbf{Time} & $\mathcal{O}(M^3 + M \cdot d)$ & $\mathcal{O}(M^2 \cdot d)$ &  $\mathcal{O}(M^2 \cdot d)$ \\
\midrule
\textbf{Space} & $\mathcal{O}(M^2 + M \cdot d)$ & $\mathcal{O}(M \cdot d)$ & $\mathcal{O}(M \cdot d)$ \\
\bottomrule
\end{tabular}
\label{tab:complexity_all_method}
\end{table}

\subsection{Time and Space Complexity Analysis}\label{method:complexity}
In this section, we analyze the computational complexity of each \method component at each time step, with results and baseline comparisons detailed in Tables~\ref{tab:complexity} and~\ref{tab:complexity_all_method}, respectively.
For analytical simplicity, we assume $d=\mathcal{O}(h)$.

\subsubsection{Memory Update.} 
Integrating a new sample involves computing cosine similarities and performing a weighted update. 
For a labeled sample, the search space is restricted to a specific class partition, yielding a time complexity of $\mathcal{O}(\frac{L}{c} \cdot h)$. 
Conversely, an unlabeled sample is compared against the entire unlabeled memory, resulting in a time complexity of $\mathcal{O}(U \cdot h)$.
Since the memory module stores only the labeled and unlabeled samples, its total space footprint is $\mathcal{O}(L \cdot d + U \cdot d) = \mathcal{O}(M \cdot d)$, where $M = L + U$.

\subsubsection{Graph-enhanced Classification via Dynamic Graph Inference.}
The embedding function $f_E$ ($\alpha_1$-layer MLP) incurs a time complexity of $\mathcal{O}(\alpha_1 \cdot M \cdot h^2)$.
The scoring function $f_S$ requires $\mathcal{O}(M^2 \cdot h)$ time due to pairwise similarity computation, and 
the pruning function $f_P$ takes $\mathcal{O}(M^2)$ time.  
Given a sparse adjacency matrix with at most $kM$ edges,
the computational cost has two components. The $\alpha_2$-layer MLP $Backbone$ model incurs a time complexity of $\mathcal{O}(\alpha_2 \cdot M \cdot h^2)$ for feature transformation, while the message passing step requires $\mathcal{O}(\alpha_2 \cdot k \cdot M \cdot h)$ time. 
Thus, the total time complexity of graph-enhanced classification via dynamic graph inference is $\mathcal{O}(M^2 \cdot h + (\alpha_1 + \alpha_2) \cdot M \cdot h^2)$.
All latent embeddings ($\mathcal{O}(M \cdot h)$) and adjacency matrices ($\mathcal{O}(M^2)$) are computed on the fly at each time step and discarded afterward, contributing no additional space overhead.  
The learnable parameters of $f_E$ require $\mathcal{O}(\alpha_1 \cdot h^2)$ space and the learnable parameters of $f_B$ require $\mathcal{O}(\alpha_2 \cdot h^2)$.

\subsubsection{Model Update.}
From the dynamic graph inference for model update, it incurs a time complexity of $\mathcal{O}(M^2 \cdot h + (\alpha_1 + \alpha_2) \cdot M \cdot h^2)$, which is the same as that of graph-enhanced classification.
In addition, the regularization term (Eq.~(\ref{eq:reg})) requires $\mathcal{O}(M^2 \cdot h)$ time to compare pairwise similarities in both the input and latent spaces.
As in graph-enhanced classification, all latent embeddings ($\mathcal{O}(M \cdot h)$) and adjacency matrices ($\mathcal{O}(M^2)$) are computed on the fly at each time step and discarded afterward, contributing no additional space overhead.  

\subsubsection{Overall Complexity of \method.}
Combining all components, the dominant computational cost per sample is
\begin{equation}\label{model:complexity}
\mathcal{O}(M^2 \cdot h  + (\alpha_1 + \alpha_2) \cdot M \cdot h^2).
\end{equation}
The total space complexity is determined by the memory module and the learnable parameters of both $f_E$ and the $Backbone$ model, yielding
\begin{equation}
\mathcal{O}(M \cdot d + (\alpha_1 + \alpha_2) \cdot h^2).
\end{equation}
Both the per-sample processing time and the space requirement remain constant with respect to the data stream size, since $M$, $h$, $d$, $\alpha_1$, and $\alpha_2$ are fixed constants, making the method well suited for streaming scenarios.

\subsection{A Lightweight Variant: \method-Light (\method-L)\label{method:light}}
While \method has a per-sample time complexity independent of the data stream size, the $M^2$ term (see Eq. (\ref{model:complexity})) may become a bottleneck as the memory buffer size $M$ grows.
To ensure scalability even with large memory buffers, we introduce \methodlight, a lightweight variant of \method. \methodlight utilizes a lightweight class-centric scoring and pruning function ($f_{S\&P}$) to reduce computational overhead from quadratic to linear in $M$.

\subsubsection{Lightweight Labeled Memory.}
To mitigate the cost associated with the number of labeled elements (i.e,  $L$), we constrain the labeled memory size to the number of classes (i.e., $c$). 
That is, exactly one prototype is maintained per class, significantly reducing memory overhead during training.

\subsubsection{Lightweight Scoring \& Pruning Function ($f_{S\&P}$).}
Rather than computing all-pair similarities for a full adjacency matrix, $f_{S\&P}$ accelerates the process by restricting weight computations to a dynamic query set. 
\begin{itemize}[leftmargin=*]
    \item \textbf{Training Phase}: Similarities are computed only between the $c$ labeled memory elements (which are queries) and all $M$ memory samples, leading to a sparse intermediate adjacency matrix $A'_t \in \mathbb{R}^{M \times M}$, where only $c \times M$ entries can be non-zero and thus computed. That is,
\begin{equation}\label{method:light_similarity} [A'_t]_{ij} = f'_S(E_t(i), E_t(j)), \quad f'_S: \mathbb{R}^h \times \mathbb{R}^h \to \mathbb{R} \end{equation}
where $i \in \{1, \dots, c\}$ and $j \in \{1, \dots, M\}$. 

\item \textbf{Inference (Classification) Phase:} The query set is reduced to the input sample $x_t$, leading to a sparse intermediate adjacency matrix $A'_t \in \mathbb{R}^{(M+1) \times (M+1)}$, where only $1 \times (M+1)$ values can be non-zero and thus computed.
\end{itemize}

To eliminate spurious connections, $A'_t$ is pruned by retaining only the top-$k$ most relevant connections per query, as follows:
\begin{equation}A_t = f_{P'}(A'_t, k).\end{equation}
This yields a sparse matrix $A_t$ containing up to $c \times k$ edges during training and up to $1 \times k$ edges during inference, drastically reducing computational overhead in both phases.

\subsubsection{Lightweight Regularization Term ($\mathcal{L}_{reg}^{L}$).}
Rather than computing all-pair similarities in the input and latent spaces for the regularization term in Eq. (\ref{eq:reg}), \methodlight focuses on the similarities between the $c$ labeled memory elements and the $M$ memory samples.

\subsubsection{Overall Complexity of \method-Light (\method-L).}
The detailed derivations are the same as those in Section~\ref{method:complexity}, except that 
the light scoring \& pruning function $f_{S\&P}$ has time complexity $\mathcal{O}(c \cdot M \cdot h)$
and the light regularization term $\mathcal{L}_{reg}^{L}$ requires $\mathcal{O}(c \cdot M \cdot h)$ time.
Consequently, the dominant computational cost per sample becomes
\begin{equation}
\mathcal{O}(c \cdot M \cdot h + (\alpha_1 + \alpha_2) \cdot M \cdot h^2),
\end{equation}
which is lower than that of \method because typically $c<<M$.
The empirical efficiency gains and the loss in accuracy are examined in the following section.
Note that the total space complexity of \methodlight remains identical to \method.
As in \method, all latent embeddings ($\mathcal{O}(M \cdot h)$) and adjacency matrices ($\mathcal{O}(c \cdot M)$) are computed on the fly at each time step and discarded afterward, contributing no additional space overhead.

\section{Experiments}

In this section, we present the experimental setup and results to examine the following research questions:
\begin{itemize}[leftmargin=*]
\item \textbf{Q1. Performance:} How effective is \method compared to baseline methods? 
\item \textbf{Q2. Ablation Study:} Do \method components contribute to performance?
\item \textbf{Q3. Scalability and Speed:} How fast and scalable is \method?
\item \textbf{Q4. Sensitivity:} How is the performance of \method affected by $L$ and $U$?
\end{itemize}

\subsection{Experiment Setup}\label{exp:setup}

\begin{table}[t!]
  \caption{Dataset Statistics}
  \label{tab:experiment_dataset}
  \centering
  \begin{tabular}{c|ccc|c|ccc}
    \toprule
    Dataset (Domain) & $|f|$ & $|c|$ & $|D|$ & Dataset (Domain) & $|f|$ & $|c|$ & $|D|$  \\
    \midrule
    MNDS (Web) & 500 & 17  & 10,917 & Shuttle (Tabular) & 7  & 7  & 58,000  \\
    Shopper (Web) & 15  & 2  & 12,330  & GSAD (Tabular) & 128  & 6  & 13,910   \\
    WebKB (Web) & 500  & 7  & 8,282 & SDD (Tabular) &  48 & 11  & 58,509 \\
    MNIST (Image) & 784  & 10  & 70,000 & HAR (Tabular) & 561  & 6  & 10,299 \\
    CIFAR-10 (Image) & 3,072  & 10  & 60,000  & OD (Tabular) & 5  & 2  & 20,560 \\
    KMNIST (Image) & 784  & 10  & 70,000  & F.MNIST (Image) & 784  & 10  &  70,000 \\
    \bottomrule
  \end{tabular}
\end{table}

\subsubsection{Datasets.}
We evaluate \method on 12 datasets: 3 web datasets, 4 image datasets, and 5 tabular datasets. 
Table~\ref{tab:experiment_dataset} summarizes the statistics of all datasets in our experiments, and  detailed dataset descriptions are as follows:

\begin{itemize}[leftmargin=*]
\item \textbf{Web:}
We use three web application datasets: MNDS~\cite{petukhova2023mn}, Shopper~\cite{shopper}, and WebKB~\cite{cmu_webkb4unis_1998}, 
extracting 500 numerical features from the text of MNDS and WebKB using TfidfVectorizer.

\item \textbf{Image:}
We use four image datasets: MNIST~\cite{deng2012mnist}, CIFAR-10~\cite{krizhevsky2009learning}, KMNIST~\cite{clanuwat2018deep}, and F.MNIST (FashionMNIST) ~\cite{FMNIST}. 
Each input sample is flattened into a vector prior to being fed into the model.

\item \textbf{Tabular:}
We use five tabular datasets from UCI Repository~\footnote{https://archive.ics.uci.edu/}: 
Shuttle (Statlog-Shuttle), GSAD (Gas Sensor Array Drift), SDD (Sensorless Drive Diagnosis), HAR (Human Activity Recognition), and OD (Occupancy Detection).
\end{itemize}

To convert the static datasets into a data-streaming format, we configure the data loader with a batch size of 1. 
Consequently, a single sample arrives at a time and is processed in a strict single-pass manner, meaning the model sees each element only once.

\subsubsection{Baselines.}
We compare \method against state-of-the-art SSL models designed for data streams: one graph-based model (TLP~\cite{wagner2018semi}) and two cluster-based models (ReSSL~\cite{din2020online} and HDSSL~\cite{din2021learning}).
To examine the contribution of each \method component, we additionally consider the following variants of \method: 

\begin{itemize}[leftmargin=*]
\item $Base$: An MLP classifier without memory buffers or graph structures.
\item +$M$: The $Base$ model with memory buffers.
\item +Static: The $Base$ model with memory buffers and a static $k$-NN graph.
\item +GSL: The proposed method but without regularization (Eq.~\eqref{eq:reg}).
\item \method: The proposed method with adaptive memories and dynamic GSL.
\end{itemize}

\subsubsection{Implementation Details.}
In our experiments, the total memory size is fixed to $M=1000$ for all methods (TLP, ReSSL, HDSSL, and \method).
Under the 1.0\% label ratio configuration, we allocate $|L|=100$ for labeled samples.
For the 0.1\% ratio, we define the labeled memory size as $|L|=\max(c, 10)$ to ensure at least one labeled sample per class, where $c$ denotes the number of classes.
In both scenarios, the remaining memory ($|U|=1000 - |L|$) is reserved for unlabeled data. 
Additional implementation details, hyperparameters, and the experimental environment are provided in Appendix A.

\subsubsection{Evaluation.}
We follow the standard test-then-train evaluation protocol commonly used for SSL on continuous data streams~\cite{ashfahani2019autonomous}. 
All reported results represent the average test accuracy over the entire duration of the data stream.
To compare the performance across different methods, we calculate the average rank by ranking each method on individual datasets and averaging these ranks.

\begin{table}[t!]
  \caption{Average accuracy over the entire data stream. The best results are in \textbf{bold}, and the second best are \underline{underlined}. \method achieves the highest accuracy for most datasets and label ratios.
  }
  \label{tab:experiment_total}
  \centering
  \begin{tabular}{c|c|ccccc}
    \toprule
    Label & Dataset & TLP & ReSSL & HDSSL & \method & \method-L \\
    \midrule
    \multirow{14}{*}{0.1\%} & MNDS & 0.0559 & 0.0809  & 0.0796   & \underline{0.2792} & \textbf{0.2974}  \\
    & Shopper & \textbf{0.8453} & 0.6304  & 0.7571  & \underline{0.8047} & 0.7510 \\
    & WebKB & 0.1980 & 0.2310  & 0.9857 & \underline{0.9943} & \textbf{0.9946}   \\
    \cmidrule{2-7}
    & MNIST  & 0.1773 & 0.3400  & 0.3797 & \textbf{0.7086} & \underline{0.6752} \\
    & CIFAR-10 & 0.1123 & 0.1064  & 0.1173   & \textbf{0.1519} & \underline{0.1486} \\
    & KMNIST & 0.2290 & 0.2293  & 0.2768   & \textbf{0.5498} & \underline{0.5190} \\
    & F.MNIST & 0.1652 & 0.3039  & 0.3918   & \underline{0.5286} & \textbf{0.5700} \\
    \cmidrule{2-7}
    & Shuttle & 0.7874 &  0.8350  & \textbf{0.8662} & 0.6914 & 0.7787 \\
    & GSAD & 0.2097 & 0.1971  & 0.2279  & \textbf{0.3237} & \underline{0.3031} \\
    & SDD & 0.7383 &  0.6827  & 0.5661   & \underline{0.9967} & \textbf{0.9984} \\
    & HAR & 0.2236 &  0.3844  & 0.2087   & \textbf{0.6083} & \underline{0.6057}  \\
    & OD & 0.7683 &  0.7717  & 0.7801 & \underline{0.7894} & \textbf{0.8136}  \\
    \cmidrule{2-7}
    & Average & 0.3759  & 0.3892 &  0.4695  & \underline{0.6189} & \textbf{0.6213}  \\
    & Avg. Rank & 4.00  & 4.00 &  3.25  & \textbf{1.83} & \underline{1.92}   \\
    \midrule
    \multirow{14}{*}{1.0\%} & MNDS & 0.1182  & 0.1915  & 0.3199  & \textbf{0.4415} & \underline{0.3929}  \\
    & Shopper & \textbf{0.8453}  & 0.7983  & 0.8025   & \underline{0.8437} & 0.6768 \\
    & WebKB & 0.4005  & 0.3223  & 0.7179   & \underline{0.9909} & \textbf{0.9967}  \\
    \cmidrule{2-7}
    & MNIST& 0.1309  & 0.6880  &  0.7842 & \textbf{0.8695} & \underline{0.8102} \\
    & CIFAR-10 & 0.1635  & 0.1503  & 0.1351  & \textbf{0.2324} & \underline{0.2126} \\
    & KMNIST  & 0.4473 & 0.5155  & 0.6464  & \textbf{0.7829} & \underline{0.6781} \\
    & F.MNIST & 0.1458  & 0.5809  & 0.6596  & \textbf{0.7126} & \underline{0.6883} \\
    \cmidrule{2-7}
   & Shuttle  &  0.7860 & \underline{0.9640}  & \textbf{0.9672} & 0.9370 & 0.8519 \\
    & GSAD  & 0.1933  & 0.6267  & 0.6718 & \textbf{0.7553} & \underline{0.6954} \\
    & SDD & 0.5497  & 0.9587  & 0.8239   & \underline{0.9951} & \textbf{0.9957} \\
    & HAR & 0.1988  & 0.6032  & 0.6682   & \textbf{0.8242} & \underline{0.7778}  \\
   & OD & 0.7728  & 0.8680  & \underline{0.8841}  & \textbf{0.8951} & 0.8673  \\
    \cmidrule{2-7}
    & Average  &  0.3960  & 0.6056 & 0.6734   & \textbf{0.7734} & \underline{0.7203}  \\
    & Avg. Rank  &  4.42  & 3.75 & 3.00   & \textbf{1.42} & \underline{2.42}  \\
    \bottomrule
  \end{tabular}
\end{table}

\subsection{Results and Discussions - Q1. Performance}
Table~\ref{tab:experiment_total} summarizes the average accuracy over 5 runs for all methods across datasets and label ratios.
\method achieves the highest accuracy, often by a substantial margin. 
Specifically, \method demonstrates an average improvement of 31.7\% and 14.8\% over the strongest baseline at 0.1\% and 1.0\% label ratios, respectively.
Notably, it outperforms the second-best baseline by 245.1\% on MNDS (0.1\% labels) and 42.1\% on CIFAR-10 (1.0\% labels).
To further evaluate whether this advantage is sustained throughout the stream, additional results for the final 20\%, 10\%, and 5\% portions of the data stream are provided in Appendix B.1. 
These results demonstrate that \method effectively leverages limited labeled data by propagating supervision through an adaptively learned graph.

\subsection{Results and Discussions - Q2. Ablation Study}
Table~\ref{tab:experiment_ablation} analyzes the contributions of each \method component. 
Adding the memory module ($+M$) 
results in the largest performance improvement across settings, highlighting that maintaining a compact history of labeled and representative unlabeled information is crucial for effective SSL on data streams.
Although adding a static $k$NN graph (+Static) improves performance in 12 out of 24 cases compared to the $Base$ results, the overall effect is modest because its fixed similarity structure cannot adapt to complex or evolving relationships in the data stream.

In contrast, \method, which dynamically learns the graph structure, achieves the highest average rank across all settings.
This improvement indicates that graph structure learning indeed allows capturing meaningful time-varying relational patterns. 
Furthermore, the performance gap between +GSL and \method highlights the necessity of regularization (Eq.~\eqref{eq:reg}) for stable dynamic GSL.

Furthermore, ablation studies on the memory update coefficients ($\alpha$ and $\beta$) verify the effectiveness of our adaptive mechanism over fixed values, with full detailed results across all datasets available in Appendix B.2.

\begin{table}[t!]
  \caption{\label{tab:experiment_ablation} Ablation study on \method components. The best results are in \textbf{bold}, and the second best are \underline{underlined}. \method with all components achieves the highest average rank, showing the benefit of dynamic graph inference. 
  }
  \centering
  \begin{tabular}{c|c|ccccc}
    \toprule 
    Label & Dataset & $Base$ & +$M$ & +Static & +GSL & \method \\
    \midrule
    \multirow{14}{*}{0.1\%} & MNDS & 0.0501  & 0.1850 & \underline{0.2698}  & 0.1923  &  \textbf{0.2792}  \\
    & Shopper & 0.7762  & 0.6839 & 0.6799  & \underline{0.8036}  &  \textbf{0.8047}  \\
    & WebKB & 0.2316  & 0.7498 & 0.6558  & \textbf{0.9943}  &  \textbf{0.9943}  \\
    \cmidrule{2-7}
    & MNIST & 0.1837  & 0.5664 & \textbf{0.7140}  & 0.5793  &  \underline{0.7086}  \\
    & CIFAR-10 & \textbf{0.1659} & \underline{0.1530} & 0.1485  & 0.1386  &  0.1519  \\
    & KMNIST & 0.0996  & 0.3521 & \underline{0.5491}  & 0.4510  &  \textbf{0.5498}  \\
    & F.MNIST & 0.1107  & \underline{0.5426} & \textbf{0.5702}  & 0.1047  &  0.5286  \\
    \cmidrule{2-7}
    & Shuttle & \underline{0.7007}  & 0.6958 & \textbf{0.7211}  & 0.6969  &  0.6914  \\
    & GSAD & 0.1984  & 0.2902 & 0.2836  & \underline{0.2966}  &  \textbf{0.3237} \\ 
    & SDD & 0.0549  & 0.6882 & 0.6825  & \textbf{0.9967}  &  \textbf{0.9967}  \\
    & HAR & 0.2568 & 0.5502 & 0.6028 & \textbf{0.6110 } &  \underline{0.6083}  \\
    & OD &  0.6897  & \textbf{0.7930} & 0.7868  & 0.7750  &  \underline{0.7894}  \\
    \cmidrule{2-7}
    & Average & 0.2932  & 0.5209 & \underline{0.5553}  & 0.5517 &  \textbf{0.6189}  \\
    & Avg. Rank & 4.17  & 3.08 & \underline{2.83}  & 2.92 &  \textbf{1.83}  \\
    \midrule
    \multirow{14}{*}{1.0\%} & MNDS & 0.1776  & 0.3608 & \underline{0.4387}  & 0.4028  &  \textbf{0.4415} \\
    & Shopper & 0.8405  & 0.8355 & 0.8124  & \underline{0.8435}  &  \textbf{0.8437} \\
    & WebKB & 0.6161  & 0.9344 & 0.9148  & \textbf{0.9909}  &  \textbf{0.9909} \\
    \cmidrule{2-7}
    & MNIST & 0.5863  & 0.7740 & \underline{0.8529}  & 0.7988  &  \textbf{0.8695} \\
    & CIFAR-10  & 0.2113  & 0.2235 & \underline{0.2299}  & 0.2267  &  \textbf{0.2324} \\
    & KMNIST & 0.4506  & 0.6386 & \underline{0.7613}  & 0.5686  &  \textbf{0.7829} \\
    & F.MNIST & 0.6307  & \textbf{0.7188} & \underline{0.7131}  & 0.6719  &  0.7126 \\
    \cmidrule{2-7}
    & Shuttle & 0.7781  & 0.9326 & \textbf{0.9384}  & \underline{0.9375}  &  0.9370 \\
    & GSAD & 0.2064  & \textbf{0.7788} & 0.7387  & 0.7457  &  \underline{0.7553} \\
    & SDD & 0.6074  & \textbf{0.9951} & 0.9945  & 0.9947  &  \textbf{0.9951} \\
    & HAR & 0.3613  & 0.7636 & \underline{0.7854}  & 0.7722  &  \textbf{0.8242} \\
    & OD & 0.7833  & \textbf{0.8968} & 0.8925  & 0.8949  &  \underline{0.8951} \\
    \cmidrule{2-7}
    & Average & 0.5208  & 0.7376 & \underline{0.7561}  & 0.7374  &  \textbf{0.7734}  \\
    & Avg. Rank & 4.83  & 2.83 & 2.83  & 2.83  &  \textbf{1.50}  \\
    \bottomrule
  \end{tabular}
\end{table}

\begin{figure}[t!]
    \centering
    \includegraphics[width=\textwidth]{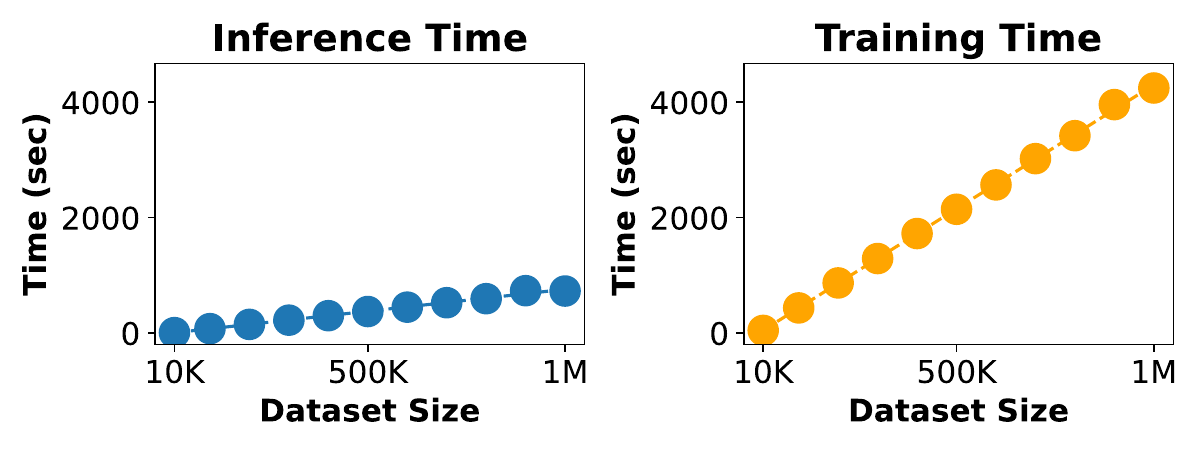} 
    \centering
    \caption{ \label{fig:scalability}
        Scalability of \method.
        Both accumulated inference time (left) and training time (right) scale linearly with the size of the input data stream, indicating a constant cost per sample.
    }
\end{figure}

\begin{table}[t!]
  \caption{Average running time of the methods. The best results are in \textbf{bold}, and the second best are \underline{underlined}.}
  \label{tab:time}
  \centering
  \begin{tabular}{c|c|ccccc}
    \toprule
    Label & Dataset  & TLP & ReSSL & HDSSL & \method & \method-L  \\
    \midrule
    \multirow{14}{*}{0.1\%} & MNDS & 108.9  & \underline{103.2}  & \textbf{32.1}  & 121.2 &  155.3   \\
    & Shopper & 105.1  & \underline{34.2}  & \textbf{23.2}  & 148.9  & 184.5  \\
    & WebKB &  \underline{74.4} &  89.3  & \textbf{26.8}  &  106.3  &  137.3   \\
    \cmidrule{2-7}
    & MNIST  &  \textbf{655.8 }&  907.7  & 1203.8  & \underline{759.7}  &  1055.8     \\
    & CIFAR-10 &  \textbf{568.7} & 3728.4   & 1928.7  & \underline{460.0}  &  681.5  \\
    & KMNIST &  \textbf{660.8} & 853.1   &  1097.2 & \underline{667.2}  &  881.3  \\
    & F.MNIST & \textbf{653.9} &  900.3  &  1347.0 &  \underline{680.3} &  953.8   \\
    \cmidrule{2-7}
    & Shuttle & 520.9 &  \underline{153.9}  & \textbf{129.7}  & 416.6 &  595.8 \\
    & GSAD &  126.3 &  \underline{67.5}  & \textbf{31.0 } &  121.3  & 165.2   \\
    & SDD & 544.6  &  \underline{212.1}   & \textbf{161.3}  & 593.8 &  818.1  \\
    & HAR & 92.9  & 106.8  &  \textbf{33.3} & \underline{90.7}   & 117.6    \\
    & OD  & 176.0  &  \underline{61.9}  & \textbf{33.0}  &  156.4  &  243.4   \\
    \cmidrule{2-7}
    & Average & \textbf{314.7} & 601.5 & 503.9 & \underline{360.2} & 499.1\\
    & Avg. Rank & \underline{2.50} & 2.75 & \textbf{2.25} & 2.92 & 4.58 \\
    \midrule
    \multirow{14}{*}{1.0\%} & MNDS  & \underline{111.9}  & 114.9  & 164.3  & \textbf{81.2} & 124.9  \\
    & Shopper & 106.7  & \underline{35.2}  & \textbf{34.5}  & 119.9  & 163.3      \\
    & WebKB & \textbf{74.8}  &  \underline{91.7}  & 158.0  &  94.4 &  125.4    \\
    \cmidrule{2-7}
    & MNIST  & \underline{685.6}  &  1017.8  & 13899.1  & \textbf{669.6}  & 879.8      \\
    & CIFAR-10 & \underline{603.9}  &  3475.8  & 12484.0 & \textbf{533.1} & 708.8        \\
    & KMNIST & \textbf{694.7} &  1077.5   & 12309.3  & \underline{825.6}  & 1012.5      \\
    & F.MNIST & \underline{689.9}  &  975.8  & 13578.1  & \textbf{632.4} & 831.7    \\
    \cmidrule{2-7}
    & Shuttle  & 538.1  &  \textbf{156.7}  &  \underline{217.4}  & 415.2  & 554.0     \\
    & GSAD & 127.5 &  \textbf{67.6}  &  139.6 &  \underline{117.4}  &  161.6    \\
    & SDD  & \underline{560.9}  &  \textbf{217.3 }  & 673.2  & 534.1  & 656.4      \\
    & HAR  & \underline{94.4}  &  115.0  & 254.4  & \textbf{90.3}   & 122.6    \\
    & OD   & \underline{181.2}  &  60.9  & \textbf{56.6}  &  172.6  & 227.1    \\
    \cmidrule{2-7}
    & Average  & \underline{372.5} & 617.2 & 4496.5 & \textbf{359.7} & 464.0 \\
    & Avg. Rank & \underline{2.33} & 2.58 & 4.00 & \textbf{2.08} & 4.00  \\
    \bottomrule
  \end{tabular}
\end{table}

\subsection{Results and Discussions - Q3. Scalability and Speed}

\subsubsection{Scalability of \method.}
We evaluate the scalability of \method using synthetic data streams ranging from 10K to 1M samples.
As shown in Figure~\ref{fig:scalability}, both inference and training times increase linearly with the stream size (i.e., constant time per sample), 
confirming its efficiency for large-scale streaming applications.

\subsubsection{Speed Comparisons.}
Table~\ref{tab:time} shows the average running time of the methods.
Although \method is the only deep learning model, it remains competitive against cluster-based methods (ReSSL, HDSSL) and even faster when $d \gg h$. 
For example, on the CIFAR-10 dataset with a $1.0\%$ label ratio, \method achieves a $95.7\%$ speedup over HDSSL.
This efficiency advantage stems from \method's extensive use of embedding-level (i.e., $h$) computation, whereas cluster-based methods rely more on feature-level (i.e., $d$) computation.

\subsubsection{\method vs. \methodlight.}
\begin{figure}[t!]
    \centering
    \includegraphics[width=\textwidth]{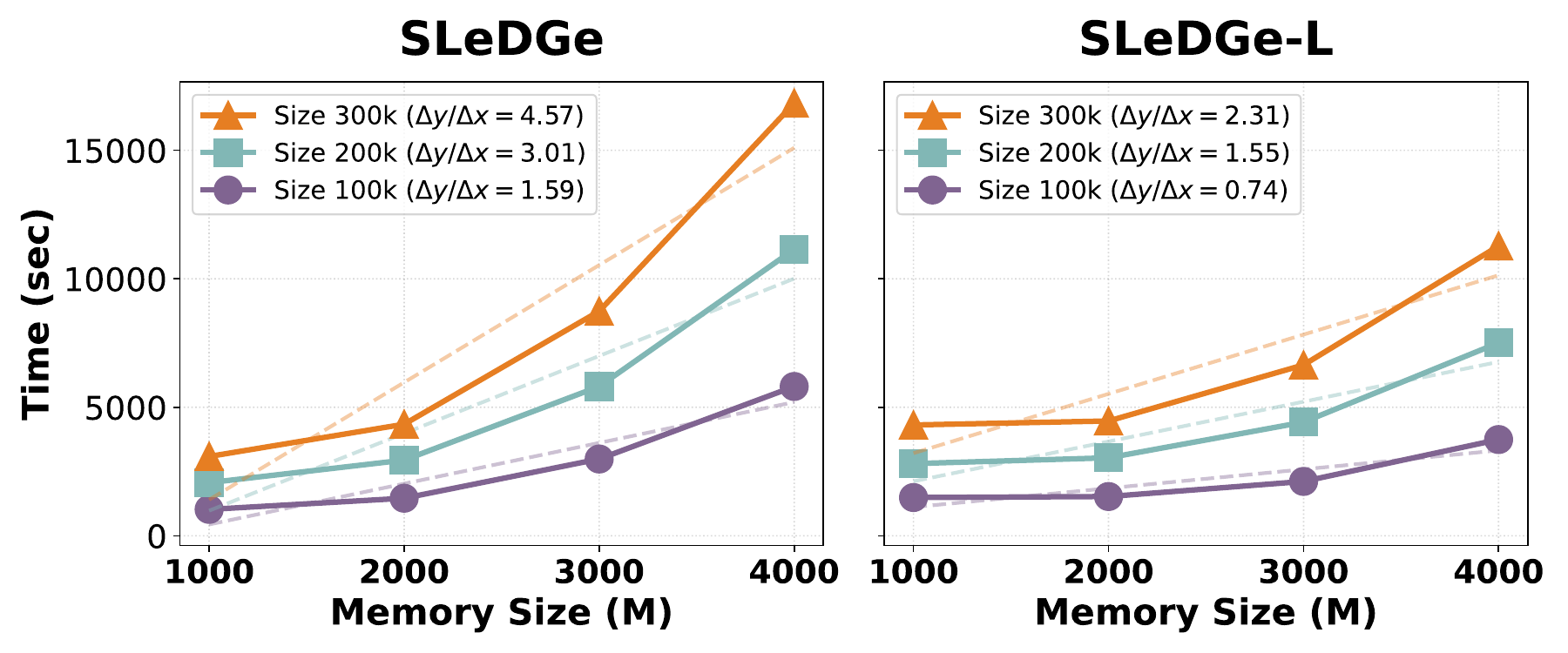}
    \centering
    \caption{ \label{fig:scalability2}
       Scalability comparison of \method and \method-L with respect to memory size.
    \methodlight exhibits significantly better scalability than \method, evidenced by the lower slopes (i.e., the rate of time increase as memory size grows).
    }
\end{figure}

We examine the execution time of \method and \methodlight while varying the memory size $M$.
As shown in Figure~\ref{fig:scalability2}, \methodlight demonstrates better scalability with $M$, consistent with our theoretical analysis.
Specifically, on a 300K-sample synthetic dataset, \methodlight maintains an average growth slope ($\Delta$Time/$\Delta M$) of 2.31, whereas \method exhibits a much steeper increase with a slope of 4.57. A notable exception occurs when $M$ is small, where the gap narrows or even reverses.
This is because the dense matrix multiplications used in the implementation of \method are highly optimized and efficiently parallelized, which can offset the algorithmic advantages of \methodlight based on sparse matrix multiplications.

\subsection{Results and Discussions - Q4. Sensitivity}
We examine the sensitivity of \method to the sizes of labeled ($L$) and unlabeled ($U$) memories. 
To isolate the impact of each component, we vary one at a time while keeping the other at its default setting.
As shown in Table~\ref{tab:memory_variants}, performance generally improves as the memory size increases.
Detailed results and analyses across all datasets are provided in Appendix B.3.

\begin{table}[t!]
  \caption{Effects of labeled ($L$) and unlabeled ($U$) memory sizes on the accuracy of \method. 
  Full results on all datasets are provided in Appendix B.3.
  }
  \label{tab:memory_variants}
  \centering
  \begin{tabular}{c|c|c|c|c|c|c}
    \toprule
     & \multicolumn{3}{c|}{$L$} & \multicolumn{3}{c}{$U$}  \\
    \midrule
    Dataset & 20 & 100 & Gain & 180 & 900 & Gain \\
    \midrule
    MNDS & 0.3904 & 0.4415 & +13.09\%  & 0.3816 & 0.4415 & +15.70\% \\
    F.MNIST & 0.6770 & 0.7126 & +5.26\%  & 0.7057 & 0.7126 & +0.98\% \\
    Shuttle & 0.7707 & 0.9370 & +21.58\%  & 0.9439 & 0.9370 & -0.73\% \\
    \bottomrule
  \end{tabular}
\end{table}

\section{Conclusions}

In this work, we present \method, a semi-supervised learning framework for data streams that integrates adaptive graph structure learning under strict memory budgets and limited labels. 
By jointly updating memories and learning adaptive graph topologies, \method effectively propagates limited supervision to new, unlabeled samples. 
Extensive experiments on 12 datasets demonstrate that \method substantially outperforms both state-of-the-art competitors and simple rule-based variants.
For reproducibility, we provide our code and datasets at \url{https://github.com/Heechan-Moon/SLeDGe}.

\section*{Acknowledgements}
This work was supported  by Institute of Information \& Communications Technology Planning \& Evaluation (IITP) grant funded by the Korea government (MSIT) (No. RS-2024-00438638, EntireDB2AI: Foundations and Software for Comprehensive Deep Representation Learning and Prediction on Entire Relational Databases, 30\%)
(No. RS-2022-II220157, Robust, Fair, Extensible Data-Centric Continual Learning, 60\%)
(No. RS-2019-II190075, Artificial Intelligence Graduate School Program (KAIST), 10\%).

\bibliographystyle{splncs04}
\bibliography{ref}

\newpage
\setcounter{page}{1}
\appendix
\section*{Appendix}
\section{Implementation Details}\label{appendix:detail_implementation}
\subsection{Experimental Environment}
We conduct all experiments with RTX A6000 GPUs (48GB VRAM), 512GB of RAM, and Intel Xeon Silver 4210R Processors.

\subsection{General Setup}
We assume no initial labeled training set.
The total memory size is fixed at 1,000 samples.
Under the 1.0\% label ratio configuration, we allocate $|L|=100$ for labeled samples.
For the 0.1\% ratio, we define the labeled memory size as $|L|=max(c, 10)$ to ensure at least one labeled sample per class, where $c$ denotes the number of classes.
In both scenarios, the remaining memory ($|U|=1000 - |L|$) is reserved for unlabeled data. 

For all deep learning models, we perform a hyperparameter grid search over the learning rate ($lr$), weight decay ($wd$), and time-decaying rate ($\tau$).
The search space for the learning rate spans $\{1e^{-5}, 5e^{-5}, ..., 1e^{-2}, 5e^{-2}\}$, while weight decay is selected from $\{0.0, 5 \times 10^{-4}\}$. The time-decaying rate $\tau$ is chosen from $\{1, \infty\}$, where $\tau = \infty$ denotes the absence of temporal weighting, assigning uniform importance to every sample regardless of its timestamp

\subsection{Baselines (Section 5.1 of the main paper)}

\begin{itemize}[leftmargin=*]
    \item \textbf{TLP}
is a graph-based semi-supervised learning method.
We adopt the labeled and unlabeled sample counts as described above and follow the algorithm specified in the original paper.

\item \textbf{ReSSL}
is a cluster-based semi-supervised learning method.
We use the authors’ official implementation with its default hyperparameters unless otherwise specified.
The memory size is set as described in our general setup.

\item \textbf{HDSSL}
is also a cluster-based semi-supervised learning method.
Since we do not have an initial labeled training set,
the Denoising Autoencoder (DAE) component could not be trained; therefore, we use only its clustering component.
We set "chunkSize" = 1 and vary the number of neighbors $k \in \{1, 5, 10\}$.
All other hyperparameters follow the original implementation.

\end{itemize}

\subsection{Model architectures and ablations \newline (Section 5.2 - Q2. Ablation Study of the main paper)}

\begin{itemize}[leftmargin=*]
    \item \textbf{Backbone ($Base$):}
Our backbone model is an $\alpha_2$-layer MLP equipped with layer normalization ($\alpha_2 \in \{1, 2, 3\}$), chosen for its flexibility. 
This $Base$ model 
operates without any memory modules.
Labeled samples update the model via single-instance training, while unlabeled samples are used only for evaluation.

\item \textbf{+$M$}:
This variant extends the $Base$ model by introducing memory retention. 
When updating the memory buffers in this configuration, we compute cosine similarity using the raw input features rather than the learned embeddings (as detailed in Section 4.1).

\item \textbf{Static}:
Building upon the +$M$ variant, this configuration incorporates a static $k$-nearest neighbors ($k$NN) graph for information propagation.
This graph is constructed by calculating cosine similarity across the static input features.
We search for 
the optimal number of neighbors $k \in \{1, 5, 10\}$.

\item \textbf{+GSL}:
To incorporate graph structure learning (GSL), we use an $\alpha_1$-layer MLP as the embedding function, where $\alpha_1 \in \{1, 2, 3\}$.
In this ablation, the dynamic GSL operates without a regularization term.
The number of neighbors $k$ is selected from $\{1,5,10\}$. 

\item \textbf{\method}:
Our proposed model utilizes the same GSL embedding function as above (an $\alpha_1$-layer MLP, $\alpha_1 \in \{1, 2, 3\}$). Unlike the +GSL variant, it includes a regularization term for dynamic GSL. 
The balancing parameter $\lambda$ is selected from \{0.001, 0.01, 0.1, 1.0\} and the number of $k \in \{1,5,10\}$.

\end{itemize}

\subsection{Hyperparameter Configurations}
We detail the specific hyperparameter settings used for \method across various datasets. 
\method relies on architectural parameters $\alpha_1$ (GSL embedding layers), $\alpha_2$ (Backbone MLP layers), the number of neighbors $k$, and the time-decaying rate $\tau$.
Unless otherwise specified, 
all experiments use the default configurations outlined in Table~\ref{tab:hyperparameters}, where paired values denote the optimal settings for label ratios of 0.001 and 0.01, respectively. 
For other hyperparameters omitted from this table (e.g., $lr, wd, \lambda$), we conduct a separate search for each experiment.

\begin{table}[t!]
\caption{Optimal hyperparameter configurations for \method.}
\label{tab:hyperparameters}
\centering
\begin{tabular}{c|c|c|c|c}
\toprule
\textbf{Dataset} & $\alpha_1 \in \{1,2,3\}$ & $\alpha_2 \in \{1,2,3\}$ & $k \in \{1,5,10\}$ & $\tau \in \{1, \infty \}$ \\
\midrule
MNDS & 2,1 & 2,1  & 5,10 & 1,1    \\
Shopper &    3,1 &        3,3        &         1,5   & $\infty$,$\infty$    \\
WebKB &      3,2          &          3,3       &        10,5   & 1,1   \\
\midrule
MNIST        &        3,3         &         2,1        &           5,5   & $\infty$,$\infty$     \\
CIFAR-10 &       1,1          &      1,2           &        10,10     & $\infty$,$\infty$     \\
KMNIST &           3,3      &     1,3            &           5,5    & $\infty$,$\infty$   \\
F.MNIST &         1,1        &      3,2           &        10,5      & $\infty$,$\infty$    \\
\midrule
Shuttle &       1,1          &      1,1           &      5,5        & $\infty$,$\infty$    \\
GSAD &         2,2        &        1,1         &         1,1       & $\infty$,$\infty$  \\
SDD &        2,2         &          2,1       &          1,1     & 1,1 \\
HAR &       2,2          &         1,1        &       10,10     & $\infty$,$\infty$     \\
OD &       1,2          &         1,1        &          1,1     & 1,1 \\
\bottomrule
\end{tabular}
\end{table}

\begin{table}[t!]
  \caption{Average accuracy over the entire data stream. The best results are in \textbf{bold}, and the second best are \underline{underlined}. \method achieves the highest accuracy for most datasets and label ratios.}
  \label{tab:experiment_total_appendix}
  \resizebox{1.0\textwidth}{!}{%
  \centering
  \begin{tabular}{c|ccccc|ccccc}
    \toprule
    Label & \multicolumn{5}{c|}{0.1\%} & \multicolumn{5}{c}{1.0\%} \\
    \midrule
    Dataset & TLP & ReSSL & HDSSL & \method & \method-L & TLP & ReSSL & HDSSL & \method & \method-L  \\
    \midrule
    \multirow{2}{*}{MNDS} & 0.0559 & 0.0809  & 0.0796   & \underline{0.2792} & \textbf{0.2974}  & 0.1182  & 0.1915  & 0.3199  & \textbf{0.4415} & \underline{0.3929}  \\
    & (0.0000) & (0.0000) & (0.0000) & (0.0138) & (0.0225) & (0.0126) & (0.0300) & (0.0250) & (0.0300) & (0.0133) \\
    \multirow{2}{*}{Shopper} & \textbf{0.8453} & 0.6304  & 0.7571  & \underline{0.8047} & 0.7510 & \textbf{0.8453}  & 0.7983  & 0.8025   & \underline{0.8437} & 0.6768 \\
    & (0.0000) & (0.1475) & (0.0554) & (0.0242) & (0.0791) & (0.0000) & (0.0317) & (0.0138) & (0.0025) & (0.0232) \\
    \multirow{2}{*}{WebKB} & 0.1980 & 0.2310  & 0.9857 & \underline{0.9943} & \textbf{0.9946}  & 0.4005  & 0.3223  & 0.7179   & \underline{0.9909} & \textbf{0.9967}  \\
    & (0.0000) & (0.0246) & (0.0064) & (0.0005) & (0.0005) & (0.0580) & (0.0368) & (0.0312) & (0.0043) & (0.0005) \\
    \midrule
    \multirow{2}{*}{MNIST}  & 0.1773 & 0.3400  & 0.3797 & \textbf{0.7086} & \underline{0.6752} & 0.1309  & 0.6880  &  0.7842 & \textbf{0.8695} & \underline{0.8102} \\
    & (0.0076) & (0.0585) & (0.0765) & (0.0109) & (0.0042) & (0.0015) & (0.0538) & (0.0080) & (0.0049) & (0.0094) \\
    \multirow{2}{*}{CIFAR-10} & 0.1123 & 0.1064  & 0.1173   & \textbf{0.1519} & \underline{0.1486} & 0.1635  & 0.1503  & 0.1351  & \textbf{0.2324} & \underline{0.2126} \\
    & (0.0034) & (0.0122) & (0.0081) & (0.0092) & (0.0014) & (0.0064) & (0.0155) & (0.0057) & (0.0057) & (0.0062) \\
    \multirow{2}{*}{KMNIST} & 0.2290 & 0.2293  & 0.2768   & \textbf{0.5498} & \underline{0.5190} & 0.4473 & 0.5155  & 0.6464  & \textbf{0.7829} & \underline{0.6781} \\
    & (0.0176) & (0.0493) & (0.0506) & (0.0148) & (0.0189) & (0.0093) & (0.0575) & (0.0372) & (0.0073) & (0.0066) \\
    \multirow{2}{*}{F.MNIST} & 0.1652 & 0.3039  & 0.3918   & \underline{0.5286} & \textbf{0.5700} & 0.1458  & 0.5809  & 0.6596  & \textbf{0.7126} & \underline{0.6883} \\
    & (0.0145) & (0.0286) & (0.0264) & (0.0277) & (0.0220) & (0.0052) & (0.0219) & (0.0088) & (0.0103) & (0.0046) \\
    \midrule
    \multirow{2}{*}{Shuttle} & 0.7874 &  0.8350  & \textbf{0.8662} & 0.6914 & 0.7787 &  0.7860 & \underline{0.9640}  & \textbf{0.9672} & 0.9370 & 0.8519 \\
    & (0.0013) & (0.0492) & (0.0140) & (0.0209) & (0.0272) & (0.0001) & (0.0030) & (0.0047) & (0.0132) & (0.0360) \\
    \multirow{2}{*}{GSAD} & 0.2097 & 0.1971  & 0.2279  & \textbf{0.3237} & \underline{0.3031} & 0.1933  & 0.6267  & 0.6718 & \textbf{0.7553} & \underline{0.6954} \\
    & (0.0192) & (0.0431) & (0.0215) & (0.0502) & (0.0534) & (0.0208) & (0.0306) & (0.0267) & (0.0121) & (0.0230) \\
    \multirow{2}{*}{SDD} & 0.7383 &  0.6827  & 0.5661   & \underline{0.9967} & \textbf{0.9984} & 0.5497  & 0.9587  & 0.8239   & \underline{0.9951} & \textbf{0.9957} \\
    & (0.0126) & (0.0592) & (0.0323) & (0.0001) & (0.0001) & (0.0444) & (0.0094) & (0.0097) & (0.0001) & (0.0025) \\
    \multirow{2}{*}{HAR} & 0.2236 &  0.3844  & 0.2087   & \textbf{0.6083} & \underline{0.6057}  & 0.1988  & 0.6032  & 0.6682   & \textbf{0.8242} & \underline{0.7778}  \\
    & (0.0250) & (0.0412) & (0.0488) & (0.0315) & (0.0664) & (0.0051) & (0.0387) & (0.0472) & (0.0090) & (0.0141) \\
    \multirow{2}{*}{OD} & 0.7683 &  0.7717  & 0.7801 & \underline{0.7894} & \textbf{0.8136}  & 0.7728  & 0.8680  & \underline{0.8841}  & \textbf{0.8951} & 0.8673  \\
    & (0.0119) & (0.0274) & (0.0127) & (0.0449) & (0.0179) & (0.0028) & (0.0146) & (0.0063) & (0.0167) & (0.0080) \\
    \midrule
    Average & 0.3759  & 0.3892 &  0.4695  & \underline{0.6189} & \textbf{0.6213}   &  0.3960  & 0.6056 & 0.6734   & \textbf{0.7734} & \underline{0.7203}  \\
    Avg. Rank & 4.00  & 4.00 &  3.25  & \textbf{1.83} & \underline{1.92}   &  4.42  & 3.75 & 3.00   & \textbf{1.42} & \underline{2.42}  \\
    \bottomrule
  \end{tabular}
  }
\end{table}

\section{Additional Experiment Results}\label{appendix:additiona_results}
\subsection{Performance over the Final Portions of the Data Stream \newline (Section 5.2 - Q1. Performance of the main paper)}
Table~\ref{tab:experiment_total_appendix} summarizes the average accuracy and the standard deviation over 5 runs for all methods across datasets and label ratios.
Furthermore, we evaluate model performance over the final 20\%, 10\%, and 5\% portions of the data stream. 
The corresponding results are presented in Table~\ref{tab:experiment_total_last_20},~\ref{tab:experiment_total_last_10} and~\ref{tab:experiment_total_last_5}, respectively. 
Overall, \method maintains superior performance across datasets and label ratios, demonstrating its ability to effectively propagate supervision and adapt to evolving data distributions throughout the stream. 

\begin{table}[t!]
  \caption{Average accuracy over the last 20\% of the data stream. The best results are in \textbf{bold}, and the second best are \underline{underlined}. \method achieves the highest accuracy for most datasets and label ratios.} 
  \label{tab:experiment_total_last_20}
  \resizebox{1.0\textwidth}{!}{%
  \centering
  \begin{tabular}{c|ccccc|ccccc}
    \toprule
    Label & \multicolumn{5}{c|}{0.1\%} & \multicolumn{5}{c}{1.0\%} \\
    \midrule
    Dataset & TLP & ReSSL & HDSSL & \method & \method-L & TLP & ReSSL & HDSSL & \method & \method-L  \\
    \midrule
    MNDS & 0.0179 & 0.0000  & 0.0000  & \textbf{0.2819} & \underline{0.2297}  & 0.0744  & 0.1289  & 0.2151  & \underline{0.3727} & \textbf{0.3997} \\
    Shopper & \textbf{0.7997} & 0.6682  & 0.6707  & \textbf{0.7997} & 0.6550  & \textbf{0.7997}  & 0.7622  & 0.7579  & \textbf{0.7997} & 0.6510 \\
    WebKB & 0.0000 & 0.0302  & 0.7934  & \underline{0.9925} & \textbf{0.9948}  & 0.3650  & 0.2001  & 0.4124  & \underline{0.9804} & \textbf{0.9960} \\
    \midrule
    MNIST & 0.1648 & 0.4339  & 0.5102  & \textbf{0.7952} & \underline{0.7523}  &  0.1248 & 0.7826  & \underline{0.8964}  & \textbf{0.9177} & 0.8883 \\
    CIFAR-10 & 0.1144 & 0.1137  & 0.1183  & \underline{0.1623} & \textbf{0.1656}  &  0.1645 &  0.1704 & 0.1210  & \textbf{0.2595} & \underline{0.2398} \\
    KMNIST & 0.1895 & 0.2642  &  0.2668  & \textbf{0.5182} & \underline{0.4940}  & 0.4072  & 0.5413  & \textbf{0.7526}  & \underline{0.7525} & 0.6651 \\
    F.MNIST & 0.1487 &  0.3853 & 0.5088  & \underline{0.5650} & \textbf{0.6144}  &  0.1484 &  0.6327 & \underline{0.7252}  & \textbf{0.7485} & 0.7128 \\
    \midrule
    Shuttle & 0.7926 & \underline{0.9176}  & \textbf{0.9374}  & 0.7485 & 0.8374  &  0.7915 & \underline{0.9798}  & \textbf{0.9868}  & 0.9392 & 0.8811  \\
    GSAD & 0.1813 & 0.1795  & 0.1557  & \textbf{0.3837} & \underline{0.3061}  & 0.1311  & 0.4825  & 0.4967  & \textbf{0.6434} & \underline{0.6247} \\
    SDD & 0.7797 & 0.7307  & 0.6639  & \underline{0.9965} & \textbf{0.9981}  & 0.9158  & 0.9761  & 0.8356  & \underline{0.9929} & \textbf{0.9960} \\
    HAR & 0.1876 & 0.4048  & 0.1769  & \textbf{0.6265} & \underline{0.6035}  &  0.1843 & 0.6646 & 0.7677  & \textbf{0.8918} & \underline{0.8329} \\
    OD & 0.7123 & 0.7404  & \underline{0.7555}  & \textbf{0.7904} & 0.7483  & 0.7332  & 0.8510  &  0.8488 & \textbf{0.8725} & \underline{0.8618} \\
    \midrule
    Average & 0.3407 & 0.4057  & 0.4631  & \textbf{0.6384} & \underline{0.6166}  &  0.4033 & 0.5977  & 0.6514 & \textbf{0.7642} & \underline{0.7291} \\
    Avg. Rank & 3.92 & 3.83  & 3.33  & \textbf{1.67} & \underline{2.08}  & 4.42  & 3.58  & 3.00  & \textbf{1.50} & \underline{2.42} \\
    \bottomrule
  \end{tabular}
  }
\end{table}

\begin{table}[t!]
  \caption{Average accuracy over the last 10\% of the data stream. The best results are in \textbf{bold}, and the second best are \underline{underlined}. \method achieves the highest accuracy for most datasets and label ratios.}
  \label{tab:experiment_total_last_10}
  \resizebox{1.0\textwidth}{!}{%
  \centering
  \begin{tabular}{c|ccccc|ccccc}
    \toprule
    Label & \multicolumn{5}{c|}{0.1\%} & \multicolumn{5}{c}{1.0\%} \\
    \midrule
    Dataset & TLP & ReSSL & HDSSL & \method & \method-L & TLP & ReSSL & HDSSL & \method & \method-L  \\
    \midrule
    MNDS & 0.0110 & 0.0000  & 0.0000  & \textbf{0.2946} & \underline{0.2814}  & 0.0744  & 0.1923  & 0.2605  & \underline{0.3786} & \textbf{0.3905} \\
    Shopper & \textbf{0.7989} & 0.6567  & 0.6448  & \textbf{0.7989} & 0.6404  & \textbf{0.7989}  & 0.7659  & 0.7494  & \textbf{0.7989} & 0.6397 \\
    WebKB & 0.0000 & 0.0043  & 0.7773  & \underline{0.9906} & \textbf{0.9937}  &  0.3324 & 0.0507  & 0.1778  & \underline{0.9768} & \textbf{0.9969} \\
    \midrule
    MNIST & 0.1583 & 0.4364  & 0.5358  & \textbf{0.8146} & \underline{0.7659}  & 0.1265  & 0.7948  & \underline{0.9065}  & \textbf{0.9284} & 0.8967 \\
    CIFAR-10 & 0.1144 & 0.1159  & 0.1167  & \underline{0.1677} & \textbf{0.1731}  & 0.1581  & 0.1696  & 0.1532  & \textbf{0.2583} & \underline{0.2343} \\
    KMNIST & 0.1768 &  0.2543 & 0.2458  & \textbf{0.4720} & \underline{0.4530}  & 0.3984  & 0.5219  & \textbf{0.7343}  & \underline{0.7215} & 0.6250 \\
    F.MNIST & 0.1424 &  0.3879 & 0.5124  & \underline{0.5710} & \textbf{0.6144}  &  0.1520 & 0.6308  & \underline{0.7243}  & \textbf{0.7473} & 0.7102 \\
    \midrule
    Shuttle & 0.7995 & \underline{0.9180}  & \textbf{0.9406}  & 0.7546 & 0.8414  & 0.7973  & \underline{0.9825}  & \textbf{0.9887}  &0.9399  & 0.8880 \\
    GSAD & 0.2495  & 0.1686  & 0.2289  & \textbf{0.3965} & \underline{0.2960}  & 0.1983  & 0.4659  & 0.4348  & \textbf{0.6407} & \underline{0.6045} \\
    SDD & 0.9137 & 0.9384  & 0.8736  & \underline{0.9965} & \textbf{0.9981}  &  \textbf{0.9998} & 0.9922  & 0.9661  & 0.9925 & \underline{0.9986} \\
    HAR & 0.1784 & 0.3934  & 0.1845  & \textbf{0.6150} & \underline{0.6058}  & 0.1732  & 0.6501  & 0.7701  & \textbf{0.8896} & \underline{0.8171} \\
    OD & 0.6926 & 0.7383  & 0.7328  & \textbf{0.8294} & \underline{0.7675}  &0.7344   & \textbf{0.8980}  & 0.8892  & 0.8811 & \underline{0.8903} \\
    \midrule
    Average & 0.3530 & 0.4177  & 0.4828  & \textbf{0.6418} & \underline{0.6192}  & 0.4120  & 0.5929  & 0.6462  & \textbf{0.7628} & \underline{0.7243} \\
    Avg. Rank & 4.17 & 3.50  & 3.50  & \textbf{1.67} & \underline{2.00}  &  4.08 & 3.42  &  3.08 & \textbf{1.83} & \underline{2.50} \\
    \bottomrule
  \end{tabular}
  }
\end{table}

\begin{table}[t!]
  \caption{Average accuracy over the last 5\% of the data stream. The best results are in \textbf{bold}, and the second best are \underline{underlined}. \method achieves the highest accuracy for most datasets and label ratios.} 
  \label{tab:experiment_total_last_5}
  \resizebox{1.0\textwidth}{!}{%
  \centering
  \begin{tabular}{c|ccccc|ccccc}
    \toprule
    Label & \multicolumn{5}{c|}{0.1\%} & \multicolumn{5}{c}{1.0\%} \\
    \midrule
    Dataset & TLP & ReSSL & HDSSL & \method & \method-L & TLP & ReSSL & HDSSL & \method & \method-L  \\
    \midrule
    MNDS & 0.0055 & 0.0000  & 0.0000  & \textbf{0.3949} & \underline{0.3336}  & 0.0976  & 0.1872  & 0.2804  & \underline{0.4125} & \textbf{0.4209} \\
    Shopper & \textbf{0.8214} & 0.6711  & 0.6601  & \textbf{0.8214} & 0.6455  & \textbf{0.8214}  & 0.7841  & 0.7591  & \textbf{0.8214} & 0.6662 \\
    WebKB & 0.0000 & 0.0087  & 0.7546  & \underline{0.9812} & \textbf{0.9874} & 0.2937  & 0.0696  & 0.1952  & \underline{0.9715} & \textbf{0.9937} \\
    \midrule
    MNIST & 0.1600 & 0.4222  & 0.5516  & \textbf{0.8356} & \underline{0.7891}  & 0.1274  & 0.8054  & \underline{0.9274}  & \textbf{0.9502} & 0.9180 \\
    CIFAR-10 & 0.1121 &  0.1120 &  0.1101 & \underline{0.1702} & \textbf{0.1750}  & 0.1562  & 0.1695  &  0.1551 & \textbf{0.2549} & \underline{0.2315} \\
    KMNIST & 0.1740 & 0.2554  & 0.2495  & \textbf{0.4711} & \underline{0.4566}  & 0.4000  & 0.5359  & \textbf{0.7402}  & \underline{0.7243} & 0.6430 \\
    F.MNIST & 0.1462 &  0.3901 & 0.5162  & \underline{0.5745} & \textbf{0.6194}  &  0.1498 &  0.6337 & \underline{0.7261}  & \textbf{0.7522} & 0.7127 \\
    \midrule
    Shuttle & 0.7997 & \underline{0.9239}  & \textbf{0.9422}  & 0.7659 & 0.8515  & 0.7953  & \underline{0.9862}  & \textbf{0.9898}  & 0.9441 & 0.8950 \\
    GSAD & \underline{0.3281} &  0.1528 & 0.2414  & \textbf{0.4063} & 0.3258  &  0.1727 & 0.5951  &  0.4561 & \textbf{0.7447} & \underline{0.6521} \\
    SDD & 0.9154 & \textbf{1.0000}  &  0.9488 & \textbf{1.0000} &\textbf{1.0000}  &  \textbf{1.0000} & \textbf{1.0000}  & \textbf{1.0000}  & \textbf{1.0000} & \textbf{1.0000} \\
    HAR & 0.1611 & 0.3615  & 0.1911  & \underline{0.5782} & \textbf{0.5887}  & 0.1455  & 0.6844  & 0.7899  & \textbf{0.9187} & \underline{0.8518} \\
    OD & 0.7967 & 0.8156  & 0.8189  & \textbf{0.8994} & \underline{0.8716}  & 0.8804  & 0.9198  & \textbf{0.9649}  & \underline{0.9212} & 0.9068 \\
    \midrule
    Average & 0.3684 & 0.4261  & 0.4987  & \textbf{0.6582} & \underline{0.6370}  & 0.4200  & 0.6142  & 0.6654  & \textbf{0.7846} & \underline{0.7410} \\
    Avg. Rank & 4.00 & 3.42  & 3.50  & \textbf{1.67} & \underline{2.00}  &  4.08 & 3.33  & \underline{2.58}  & \textbf{1.50} & \underline{2.58} \\
    \bottomrule
  \end{tabular}
  }
\end{table}

\subsection{Memory Update Coefficient Analysis \newline (Section 5.2 - Q2. Ablation Study of the main paper)}
We investigate the effect of the memory update coefficients, $\alpha$ and $\beta$, on model performance. 
To validate the effectiveness of our adaptive update mechanism, we compare the proposed dynamically computed coefficients against fixed values. 
As shown in Table~\ref{tab:memory_ablation_appendix}, 
applying a high coefficient (e.g., 0.95) generally yields better performance compared to a low coefficient (e.g., 0.05), demonstrating superiority in 10 out of 12 datasets. 
However, unconditionally employing a high coefficient often leads to suboptimal results. 
For instance, on the CIFAR-10 dataset, our proposed method achieves 0.2471, noticeably outperforming the 0.2339 obtained using a fixed high coefficient.
These results confirm that our update mechanism is appropriately adaptive, offering a more reasonable approach than unconditionally using a high fixed value.

\begin{table}[t!]
  \caption{Effects of fixed versus dynamically computed memory update coefficients ($\alpha$ and $\beta$) on the accuracy of \method when label ratio is 1.0\%. The proposed adaptive approach successfully ensures optimal accuracy, often outperforming the solid performance of a high fixed value.
  }
  \label{tab:memory_ablation_appendix}
  \centering
  \begin{tabular}{c|c|c|c}
    \toprule
    Coefficient & \multicolumn{3}{c}{$\alpha, \beta$} \\
    \midrule
    Dataset & 0.05 & 0.95 & Proposed \\
    \midrule
     MNDS & 0.3905 & 0.4080 & 0.3907 \\
    Shopper & 0.8392 & 0.8385 & 0.8391 \\
    WebKB & 0.9593 & 0.9628 & 0.9641 \\
    \midrule
    MNIST & 0.7270 & 0.8332 & 0.7758 \\
    CIFAR-10 & 0.2315 & 0.2339 & 0.2471  \\
    KMNIST & 0.5461 & 0.6412 & 0.6132 \\
    F.MNIST & 0.6871 & 0.7256 & 0.7114 \\
    \midrule
    Shuttle & 0.9415 & 0.9372 & 0.9403  \\
    GSAD & 0.7978 & 0.8222 & 0.8104 \\
     SDD & 0.9931 & 0.9957 & 0.9948  \\
    HAR & 0.7939 & 0.7939 & 0.7960 \\
     OD & 0.8885 & 0.9022 & 0.8994 \\
    \bottomrule
  \end{tabular}
\end{table}

\subsection{Memory Size Analysis \newline (Section 5.2 - Q4. Sensitivity of the main paper)}
We evaluate the model's performance by independently varying the labeled memory size ($L$) and the unlabeled memory size ($U$).
To isolate the specific impact of each component, we perform a sensitivity analysis by varying 
only one size ($L$ or $U$) at a time, keeping all other hyperparameters constant.
Our baseline configuration uses $L=100$ and $U=900$ (reflecting the 1.0\% label ratio setting).
We test variations by reducing these values to 20\% of their original levels (e.g., $L = 20$ or $U = 180$).

Table~\ref{tab:memory_variants_appendix} shows that performance generally improves as memory size is expanded.
Specifically, increasing $L$ leads to performance enhancements across 11 out of 12 datasets, achieving an average relative gain of 5.24\%.
In contrast, increasing $U$ yields improvements in 8 out of 12 datasets, resulting in a more modest average gain of 2.41\%.
These results indicate that \method exhibits higher sensitivity to the availability of labeled samples than to unlabeled data.

\begin{table}[t!]
  \caption{\method results across varying labeled ($L$) and unlabeled ($U$) memory sizes. 
  Increasing either memory size generally leads to performance gains.
  }
  \label{tab:memory_variants_appendix}
  \centering
  \begin{tabular}{c|c|c|c|c|c|c}
    \toprule
    Memory & \multicolumn{3}{c|}{$L$} & \multicolumn{3}{c}{$U$}  \\
    \midrule
    Dataset & 20 & 100 & Gain & 180 & 900 & Gain \\
    \midrule
     MNDS & 0.3904 & 0.4415 & + 13.09\% & 0.3816 & 0.4415 & + 15.70\% \\
    Shopper & 0.8405 & 0.8437 & + 0.38\% & 0.8434 & 0.8437 & + 0.04\% \\
    WebKB & 0.9922 & 0.9909 & - 0.13\% & 0.9925 & 0.9909 & - 0.16\% \\
    \midrule
    MNIST & 0.8480 & 0.8695 & + 2.54\% & 0.8291 & 0.8695 & + 4.87\% \\
    CIFAR-10 & 0.2163 & 0.2324 & + 7.44\% & 0.2342 & 0.2324 & - 0.77\% \\
    KMNIST & 0.7290 & 0.7829 & + 7.39\% & 0.7325 & 0.7829 & + 6.88\% \\
    F.MNIST & 0.6770 & 0.7126 & + 5.26\% & 0.7057 & 0.7126 & + 0.98\% \\
    \midrule
    Shuttle & 0.7707 & 0.9370 & + 21.58\% & 0.9439 & 0.9370 & - 0.73\% \\
    GSAD & 0.7289 & 0.7553 & + 3.62\% & 0.7610 & 0.7553 & - 0.75\% \\
     SDD & 0.9946 & 0.9951 & + 0.05\% & 0.9944 & 0.9951 & + 0.07\% \\
    HAR & 0.8140 & 0.8242 & + 1.25\% & 0.8051 & 0.8242 & + 2.37\% \\
     OD & 0.8913 & 0.8951 & + 0.43\% & 0.8914 & 0.8951 & + 0.42\% \\
    \bottomrule
  \end{tabular}
\end{table}

\end{document}